\documentclass{article}

\usepackage{primearxiv}

\usepackage[utf8]{inputenc} 
\usepackage[T1]{fontenc}    
\usepackage{hyperref}       
\usepackage{url}            
\usepackage{booktabs}       
\usepackage{amsfonts}       
\usepackage{nicefrac}       
\usepackage{microtype}      
\usepackage{lipsum}
\usepackage{fancyhdr}       
\usepackage{graphicx}       

\usepackage{xcolor}
\usepackage{todonotes}
\usepackage{amsmath}
\usepackage{amsthm}
\usepackage{adjustbox}

\usepackage{subcaption}
\usepackage{caption} 
\usepackage{array}
\usepackage{colortbl}

\usepackage{listings}
\usepackage{xcolor}       
\usepackage{multirow}
\usepackage{tcolorbox}
\usepackage{tikz}
\usetikzlibrary{decorations.pathreplacing}

\colorlet{lightyellow}{yellow!40}

\makeatletter
{\small 
\xdef\f@size@small{\f@size}
\xdef\f@baselineskip@small{\f@baselineskip}
\normalsize 
\xdef\f@size@normalsize{\f@size}
\xdef\f@baselineskip@normalsize{\f@baselineskip}
}
\newcommand{\smalltonormalsize}{%
  \fontsize
    {\fpeval{(\f@size@small+\f@size@normalsize)/2}}
    {\fpeval{(\f@baselineskip@small+\f@baselineskip@normalsize)/2}}%
  \selectfont
}
\makeatother

\usepackage{amsmath, amssymb, amsthm}
\usepackage{mathtools} 

\theoremstyle{remark}

\title{Do LLMs Make Mistakes Like Students? \\ Exploring Natural Alignment between \\ Language Models and Human Error Patterns}

\author{
  Naiming Liu, \, Shashank Sonkar, \, Richard G. Baraniuk \\
  Rice University \\
  Houston, TX \\
  \texttt{nl35@rice.edu} \\
}

\begin{document}
\maketitle

\begin{abstract}
Large Language Models (LLMs) have demonstrated remarkable capabilities in various educational tasks, yet their alignment with human learning patterns, particularly in predicting which incorrect options students are most likely to select in multiple-choice questions (MCQs), remains underexplored. Our work investigates the relationship between LLM generation likelihood and student response distributions in MCQs with a specific focus on distractor selections. We collect a comprehensive dataset of MCQs with real-world student response distributions to explore two fundamental research questions: (1). RQ1 - Do the distractors that students more frequently select correspond to those that LLMs assign higher generation likelihood to? (2). RQ2 - When an LLM selects a incorrect choice, does it choose the same distractor that most students pick? Our experiments reveals moderate correlations between LLM-assigned probabilities and student selection patterns for distractors in MCQs. Additionally, when LLMs make mistakes, they are more likley to select the same incorrect answers that commonly mislead students, which is a pattern consistent across both small and large language models. Our work provides empirical evidence that despite LLMs' strong performance on generating educational content, there remains a gap between LLM's underlying reasoning process and human cognitive processes in identifying confusing distractors. Our findings also have significant implications for educational assessment development. The smaller language models could be efficiently utilized for automated distractor generation as they demonstrate similar patterns in identifying confusing answer choices as larger language models. This observed alignment between LLMs and student misconception patterns opens new opportunities for generating high-quality distractors that complement traditional human-designed distractors.
\end{abstract}


\section{Introduction}
The emergence of Large Language Models (LLMs) has  demonstrated remarkable capabilities across diverse educational applications: from generating curriculum materials \cite{moein2024beyond} to providing personalized tutoring \cite{alsafari2024towards,sonkar2023class,schmucker2024ruffle} and real-time feedback \cite{nicolicioiu2024panza,bewersdorff2023assessing}, to modeling human cognitive processes \cite{mcintosh2024inadequacy,sonkar2024llm}. Within this broad spectrum of LLM's capabilities, a particularly intriguing direction is their potential to understand and predict patterns in student thinking, especially how students interact with multiple-choice questions and select incorrect choices (distractors).

Multiple-choice questions (MCQs) serve as powerful diagnostic tools in educational assessment \cite{eedi-mining-misconceptions-in-mathematics}. Instead of just being random wrong answers, high-quality distractors of MCQs should reflect common misconceptions or reasoning errors that students encounter. The patterns in how students select these distractors often reveal systematic misconceptions shared across learners \cite{smith1994misconceptions}, providing valuable insights into their learning processes and making MCQs particularly effective for identifying knowledge gaps in student understanding.

A fundamental question emerges: do LLMs inherently capture the patterns in how students select distractors? Understanding whether these models naturally align with student misconception patterns have significant implications for both educational practice and our understanding of LLMs' reasoning capabilities. This insight could improve how we develop educational technologies, from enhancing assessment tools to creating tutoring systems that better anticipate and address common student misunderstandings. Further, this investigation could reveal whether LLMs develop internal representations that parallel human cognitive patterns.

To investigate this question systematically, we propose an analysis framework focusing on two specific research questions:

\textbf{RQ1:} Do the distractors that students more frequently select correspond to ones that LLMs assign higher generation likelihood to?

\textbf{RQ2:}  When an LLM selects an incorrect answer, does it choose the same distractor that most commonly misleads students?

We conduct our analysis using a comprehensive dataset of $3,202$ multiple-choice questions with real-world student response distributions and two families of large language models - LLaMA \cite{touvron2023llama} and Qwen \cite{bai2023qwen}, with parameters ranging from 0.5B to 72B. To quantify the relationship between LLM generation likelihood and student response patterns in distractor selections, we introduce a novel alignment score. This study presents the first empirical investigation of whether LLMs' generation preferences and error choices align with common student misconceptions.

Our analysis reveals two key findings. First, when examining LLMs' generation probabilities for distractors (RQ1), we observe moderate correlations with student selection patterns (Pearson $r=0.28-0.37$), with larger models demonstrating slightly stronger alignment. Second, our analysis of LLMs' incorrect selections (RQ2) shows a striking tendency across both small and large models to choose the same wrong answers that students most commonly select. Even the smallest model (0.5B parameters) selects students' most common incorrect answer about $51\%$ of the time, while larger models reach up to $59\%$, indicating that this alignment with student error patterns may be an inherent property of LLMs rather than purely a function of model scale.

These findings have important implications for educational applications, particularly in MCQs design. The moderate alignment between LLM prediction and student misconception patterns suggests that while LLMs exhibit certain correlation with student thinking process, they likely approach problems differently than students do. Rather than viewing this as a limitation, it presents an opportunity: LLM-generated distractors could potentially probe different aspects of student understanding, complementing rather than replacing human expertise in the MCQs design. Notably, our finding that smaller models show similar patterns when making mistakes as larger models suggests a cost-efficient approach to generating plausible distractors. This finding points toward a promising hybrid approach where LLMs could help expand the range of assessment options by generating novel distractors while human experts ensure quality and coverage of the known misconceptions in the distractors. More broadly, our work contributes to a better understanding of how to effectively integrate LLM capabilities into educational assessment design, highlighting both their potential and limitations in modeling student cognitive process.

\section{Related Works}
\subsection{Student Misconception Analysis in Education}

Understanding and addressing student misconceptions is fundamental to improving educational outcomes, as these misconceptions arise when learners develop incorrect mental models that persist despite formal instruction \cite{chi1994conceptual}. Traditional methods for identifying misconceptions have relied on concept inventories like mal-rules \cite{malrules,sleeman}, the Force Concept Inventory in physics \cite{hestenes1992force}, while both distractors and corresponding feedback messages in math MCQs have been leveraged to help students overcome common errors through targeted remediation \cite{li2024automated,mcnichols2023automated}. Psychometric models like Item Response Theory (IRT) have been employed to measure how misconceptions influence student performance \cite{embretson2000item}, while data-driven approaches have enabled large-scale analysis of student errors, leveraging NLP and machine learning to uncover latent misconceptions in open-ended responses \cite{pardos2,pardos1}. Deep knowledge tracing models further enhance our understanding by tracking student learning trajectories and predicting future misconceptions \cite{pardos2013adapting,piech2015deep}.

Researchers have developed probabilistic models to analyze students' textual responses, aiming to uncover prevalent misconceptions without requiring instructors to predefine them \cite{michalenko2017data}, while in intelligent tutoring systems \cite{baker_test}, methods have been proposed to identify knowledge components in dialogue turns and diagnose the correctness of student responses \cite{scarlatos2023exploring}. Recent research has focused on verifying student solutions in mathematical problem-solving contexts, with emphasis on detecting errors in student reasoning and providing customized feedback \cite{daheim2024stepwise,liu2023novice}. The development of dialogue tutoring datasets with rich pedagogical properties has been explored to better understand and address student misconceptions, aiming to simulate realistic tutor-student interactions \cite{macina2023mathdial}. These various approaches and studies collectively underscore the significance of leveraging advanced computational techniques to detect and address student misconceptions, thereby informing the development of more effective educational tools and strategies.

\subsection{Automated Distractor Generation}

Researchers have investigated LLMs for automated distractor generation in MCQs \cite{baral2024automated,eedi-mining-misconceptions-in-mathematics,bitew2023distractor}, with studies revealing that while LLMs can produce mathematically valid distractors, they often fall short in anticipating common student errors or misconceptions \cite{feng2024exploring}. This limitation has prompted the development of enhanced approaches, including model training pipelines incorporating pairwise rankers to assess distractor plausibility based on potential student misconceptions. Using synthetic student choice datasets, researchers have also demonstrated improved generation of distractors that align more closely with common student misunderstandings \cite{lee2025generating}. Similarly, an overgenerate-and-rank approach has been proposed that trains ranking models to predict how likely distractors are to be selected by real students \cite{scarlatos2024improving}.

In-context learning techniques have emerged as another promising direction for generating both distractors and corresponding feedback messages in math MCQs. By providing LLMs with examples of questions and corresponding distractors, these techniques guide the generation process toward more relevant outputs \cite{mcnichols2023automated}. These collective advancements in automated distractor generation aim to reduce the manual effort required in crafting assessments while improving their diagnostic capabilities through closer alignment between distractors and actual student misconceptions.

\section{Dual Analysis Framework for LLM-Student Misconception Alignment}
To investigate whether LLMs naturally capture student misconception patterns, we analyze their behavior on multiple-choice questions and compare it against actual student response data. Our investigation focuses on two key aspects: (1). whether LLMs assign higher probabilities to incorrect answers that commonly mislead students, and (2). whether LLMs tend to select the same wrong answers that students frequently choose. To enable this analysis, we develop a systematic framework as follows:

\subsection{Preliminaries}

Let $\mathcal{Q} = \{q_1, q_2, \ldots, q_n\}$ denote our collection of $n$ multiple-choice questions. For each question $q_i$, we define:

\begin{itemize}
    \item A set of $m$ answer options $\mathcal{A}_i = \{a_{i1}, a_{i2}, \cdots, a_{im}\}$
    \item The correct answer index $c_i \in \{1, 2, \cdots, m\}$
    \item The empirical student response distribution $\mathcal{S}_i = \{s_{i1}, s_{i2}, \cdots , s_{im}\}$, where $s_{ij}$ represents the proportion of students selecting option $j$ for question $i$ such that $\sum_{j=1}^m s_{ij} = 1$
\end{itemize}

This formalization enables our analysis of the alignment between LLM behavior and student misconception patterns through both probabilistic correlation measures and direct incorrect answer comparisons.

\subsection{Likelihood Calculation of LLM's Answer Choice}

We quantify an LLM's preference for each answer choice using two formatting approaches implemented through the EleutherAI Harness evaluation framework~\cite{eval-harness}.

\subsubsection{Index-based Approach:}
In the index-based approach, we present the model with both the question and all answer choices in a structured format, where each answer choice is assigned a corresponding letter index (A, B, C, ...). The model then predicts a single letter choice representing the answer choice. For choice index $j \in \{A, B, C, \cdots\}$ of question $q_i$, we compute the log-likelihood as:

$$\mathcal{L}_{a_{ij}}^{\text{index}} = \log P(j \mid q_i, \mathcal{A}_i; \theta)$$

where $\theta$ represents the model parameters.

\subsubsection{Text-based Approach:}
In the text-based approach, we evaluate each answer choice independently by computing its likelihood when paired with only the question as input. For each choice $a_{ij}$ from question $q_i$, we compute the log-likelihood by summing over all tokens in the answer text and take an average to avoid giving long answers an unfair advantage:

$$\mathcal{L}_{a_{ij}}^{\text{text}} = \frac{\sum_{t=1}^{T_{ij}} \log P(x_t^{(ij)} \mid q_i, x_{<t}^{(ij)}; \theta)}{T_{ij}}$$

where $x_t^{(ij)}$ is the $t$-th token in choice $a_{ij}$; $T_{ij}$ is the number of tokens in choice $a_ij$ and $x_{<t}^{(ij)}$ represents all preceding tokens in the answer. The summation and average is necessary here because we measure the likelihood of the entire answer text sequence, token by token. 

The two approaches differ in their computation of answer probabilities. Index-based approach computes the conditional probability of each index given the complete context of all options, while text-based approach evaluates the likelihood of each answer text independently through token-wise probability estimation.

\subsubsection{Probability Normalization:}
For both approaches, we convert log-likelihoods to probabilities using the softmax function:

$$P_{ij}(a_{ij} \mid q_i) = \frac{\exp(\mathcal{L}_{a_{ij}})}{\sum_{j=1}^{n} \exp(\mathcal{L}_{a_{ij}})},$$

where $\mathcal{L}_{a_{ij}}$ represents either $\mathcal{L}_{a_{ij}}^{\text{index}}$ or $\mathcal{L}_{a_{ij}}^{\text{text}}$ depending on the approach used.

\subsection{Analysis Framework}

Our investigation employs two approaches to understand how LLMs align with student misconceptions and answer the proposed research questions.

\subsubsection{RQ1: Correlation Between LLM Generation Probabilities and Student Selection Patterns for Distractors:}

For each question-option pair, we compute the LLM's predicted likelihood distribution $P_{ij}$ and obtain student selection probability $s_{ij}$ for each option $a_{ij}$ over all answers choices $\mathcal{A}_i$ with $c_i$ being the correct answer. Then, we adopt statistical methods to compute the correlation coefficient $\rho_i$ between the language model's generation likelihood and student selection frequencies:

\begin{equation}
    \rho_i = \text{corr}(\{P_{ij, j \neq c_i}\}, \{s_{ij, j \neq c_i}\})
\end{equation}

By comparing the LLM's likelihood patterns with actual student selection rates, we can assess whether the model's confidence in distractors aligns with the cognitive misconceptions that commonly mislead students. This alignment, if present, would suggest that LLMs may naturally capture aspects of human cognitive biases and misconceptions that lead to systematic errors in problem-solving.

Our analysis specifically excludes correct answers $(j \neq c_i)$ to isolate these misconception patterns. By focusing only on incorrect options, we can better examine the distribution of plausible but incorrect reasoning, rather than having our analysis confounded by the model's ability to identify correct answers. Higher likelihood scores indicate higher model confidence in an answer choice, allowing us to quantify which incorrect options the model finds most plausible and compare this directly with student error patterns.

\subsubsection{RQ2: Alignment of LLM Mistakes and Student Misconception Patterns:}
Our second analysis examines whether LLM errors align with the most common student misconceptions. Specifically, we investigate if an LLM, when selecting an incorrect answer, chooses the same incorrect option that most students select. This behavioral analysis complements our perplexity-based approach by examining the model's actual answer selections rather than its probability distributions. The analysis proceeds as follows:

\begin{enumerate}
    \item For each question $q_i$, we record the LLM's selected choice option $o_i$ and define an error indicator:
    \begin{equation}
        e_i = \begin{cases}
            1 & \text{if } o_i \neq c_i \\
            0 & \text{otherwise}
        \end{cases}
    \end{equation}

    \item For cases where $e_i = 1$, we compute an alignment score:
    \begin{equation}
        \alpha_i = \frac{s_{i,o_i}}{\max\limits_{j \neq c_i} s_{ij}}
    \end{equation}
\end{enumerate}

The alignment score $\alpha_i$ quantifies how well the LLM's incorrect answers correspond to the most common student misconceptions. A score of $\alpha_i = 1$ indicates perfect alignment, meaning the LLM selected the same incorrect option that students most frequently chose when making mistakes.
To illustrate this measure, consider a multiple-choice question where students' selection rates for incorrect options are distributed as follows: Option B (30\%), Option C (20\%), and Option D (15\%). In this case, Option B represents the most common student misconception. If an LLM makes an error by selecting Option C, the alignment score would be calculated as $\alpha = \frac{20\%}{30\%} = 0.67$. This score reflects that while the LLM did select a relatively common incorrect answer, it did not align with the predominant student misconception (Option B). Lower alignment scores indicate greater divergence between the LLM's error patterns and the predominant student misconceptions. This metric provides valuable insights into whether LLMs naturally capture the cognitive biases that lead students toward specific incorrect answers, helping us understand the extent to which these models might reflect human-like patterns of misunderstanding. The mean alignment score across all questions is defined as:

$$\bar{\alpha} = \frac{\sum_{i=1}^N \alpha_i e_i}{\sum_{i=1}^N e_i}$$

\section{Experiments and Results}
\begin{table*}[t]
\centering
\caption{
Our dataset contains $3,202$ Multiple-Choice questions collected from three Major Educational Assessment Sources. The dataset spans six subject areas, with Biology and Mathematics having the largest representation. \textit{Avg. Correctness} shows the average percentage of students who selected the correct answer across all questions in each subject.
}
\scalebox{0.98}{
\begin{tabular}{lcc}
\toprule
\textbf{Subject Area} & \textbf{Num of Questions} & \textbf{Avg. Correctness (\%)} \\
\midrule
Mathematics & 807 & 58.9 \\
Biology & 946 & 60.0 \\
Physics & 636 & 63.3 \\
Social Sciences & 260 & 61.9 \\
Reading Comprehension & 285 & 56.1 \\
Humanities & 268 & 63.8 \\
\midrule
Total & 3202 & 60.5 \\
\bottomrule
\end{tabular}
}
\label{tab:dataset}
\end{table*}


In this section, we present a comprehensive empirical analysis examining how well LLMs capture student misconception patterns in multiple-choice questions. Through two complementary analyses - one focusing on LLMs' generation probabilities and another on their actual answer selections - we investigate whether these models naturally encode patterns in how students select incorrect answers.

\subsection{Student Performance Dataset}

Our analysis uses a dataset of $3,202$ multiple-choice questions drawn from six core academic domains: mathematics, biology, physics, social science, reading comprehension, and humanities. 
These questions were collected from three established educational assessment platforms, with detailed subject distribution shown in Table~\ref{tab:dataset}. To ensure reliable analysis of student performance patterns, we applied two filtering criteria: (1) each question must have responses from at least 50 students, and (2) the error rate must exceed 5\% to enable meaningful analysis of misconceptions. For consistency in our analysis, we included only questions with exactly four answer choices. The aggregated student performance data shows an average correct response rate of 60.5\% across all subjects.

\begin{table}[t]
\centering
\caption{Correlation between student selection frequencies and LLM generation probabilities for incorrect answer choices in multiple-choice questions. We adopt both index-based and text-based approach to obtain generation probabilities and report Pearson, Spearman, and Kendall correlation coefficients.}
\label{tab:rq1}
\adjustbox{max width = \textwidth}{
\begin{minipage}{0.55\textwidth}
\centering
\begin{tabular}{lccc}
\toprule
\rowcolor{gray!21} \multicolumn{4}{c}{Index-based Likelihood} \\
 & \textbf{Pearson $\uparrow$} & \textbf{Spearman $\uparrow$} & \textbf{Kendall$\uparrow$} \\
\midrule
\textbf{qwen-7b} & 0.341 & 0.331 & 0.301  \\
\textbf{llama-8b} & 0.283 & 0.269 & 0.240  \\
\textbf{qwen-14b} & \textbf{0.365} & \textbf{0.348} & \textbf{0.319}  \\
\textbf{llama-70b} & 0.296 & 0.277 & 0.269 \\
\textbf{qwen-72b} & 0.352 & 0.332 & 0.302  \\
\bottomrule
\end{tabular}
\end{minipage}
\hspace{0.08\textwidth}
\begin{minipage}{0.55\textwidth}
\centering
\begin{tabular}{ccc}
\toprule
\rowcolor{gray!21} \multicolumn{3}{c}{Text-based Likelihood} \\
\textbf{Pearson $\uparrow$} & \textbf{Spearman $\uparrow$} & \textbf{Kendall $\uparrow$} \\
\midrule
 0.131 & 0.130 & 0.134  \\
 0.149 & 0.141 & 0.127  \\
 0.162 & 0.151 & 0.139  \\
 0.176 & \textbf{0.176} & \textbf{0.160} \\
 \textbf{0.183} & 0.170 &  0.154 \\
\bottomrule
\end{tabular}
\end{minipage}
}
\end{table}


\begin{figure}[t!]
    \centering
        \includegraphics[width=0.8\textwidth]{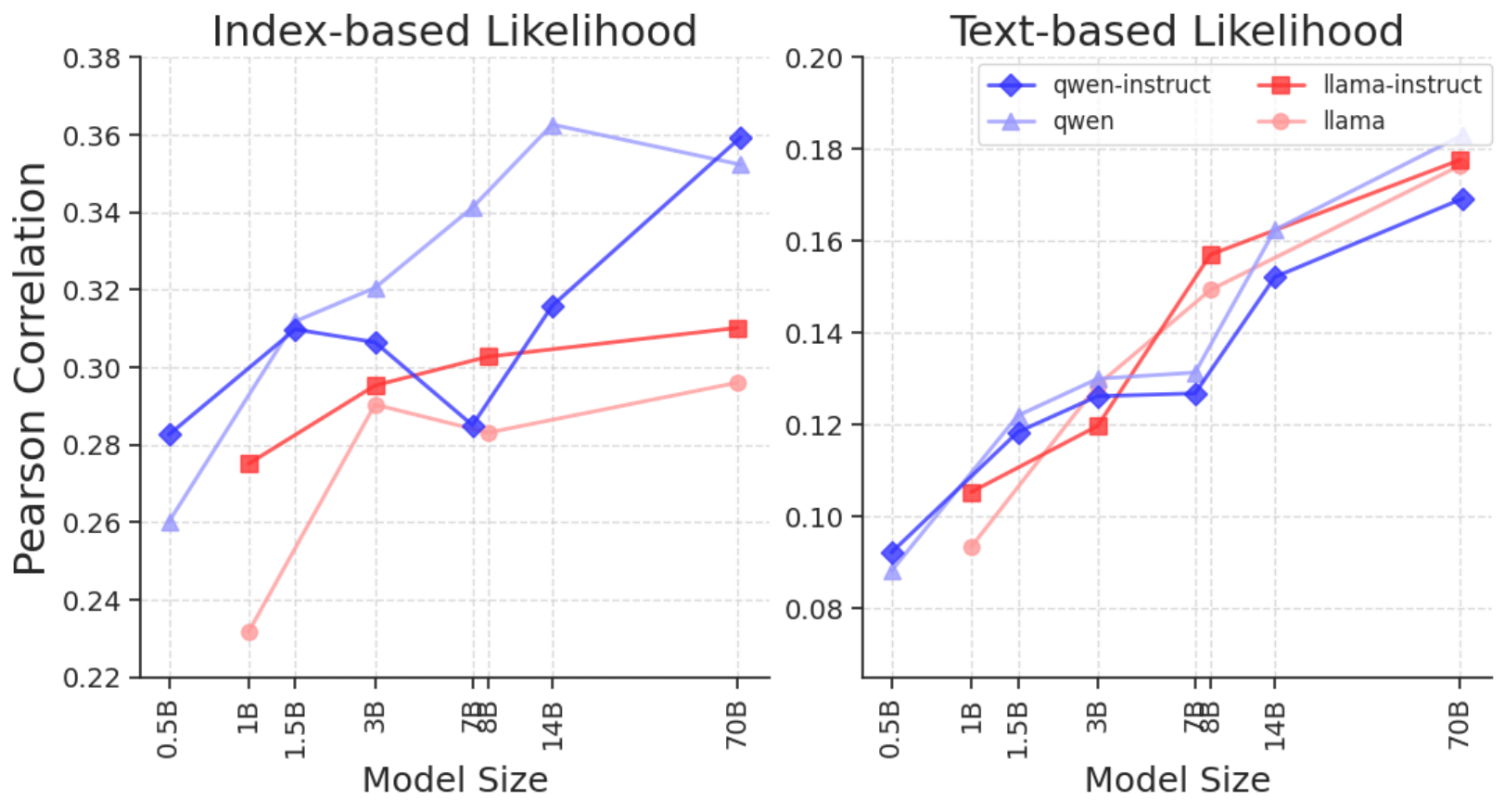} 
    \caption{
    Pearson correlation between LLM generation probabilities and student selection frequencies for incorrect answer choices (distractors) across model sizes. The index-based approach (left) measures correlation for A/B/C/D label selection probabilities, while the text-based approach (right) measures correlation for full distractor text generation probabilities. Results shown for base and instruction-tuned variants of LLaMA and Qwen model families demonstrate relatively stronger alignment between LLMs and student distractor selection patterns as model size increases, especially for text-based approach.
    }
    \label{img:rq1_alignment_size}
\end{figure}

\subsection{RQ1: Correlation Between LLM Generation Probabilities and Student Selection for Distractors}

Our first research question examines whether LLMs assign higher generation probabilities to the same incorrect answers that students commonly select. To investigate this systematically, we analyzed statistical correlations between LLM generation probabilities and student selection frequencies with Pearson's Correlation, Spearman's Rank Correlation and Kendall's tau Correlation (Details can be found in Appendix~\ref{app:stats-corr}). We tested this using two distinct probability measurement approaches: index-based, where models assign probabilities to answer labels (A/B/C/D), and text-based, where models compute generation probabilities for complete answer texts across different model sizes (0.5B to 72B parameters) and architectures (Qwen and LLaMA families\footnote{We use Qwen-2.5 for all parameter sizes; LLaMA-3.1 for 8b and 70b and LLaMA-3.2 for 1b and 3b parameter sizes.}). As shown in Table \ref{tab:rq1} and Figure \ref{img:rq1_alignment_size}, our analysis reveals several key patterns.

\subsubsection{Difference on Likelihood Approach:} The index-based approach shows better alignment between LLM likelihood and student selection distribution on distractors, with correlations ranging from 0.28 to 0.37 (Pearson). Most notably, qwen-14b achieves a correlation of 0.365 (Pearson) and 0.348 (Spearman) with student selection patterns - a moderate correlation given that these models were never explicitly trained to capture student misconceptions. This moderate correlation suggests that LLMs inherently encode some meaningful patterns about how students reason through multiple-choice questions, but are still not able to fully understand the cognitive processes underlying student misconceptions. Such natural alignment between LLM probabilities and student distractor preferences opens exciting possibilities for using these models to understand and predict student misconceptions.

The index-based approach consistently shows stronger correlations compared to text-based across all models and correlation metrics. This substantial difference (e.g., Qwen-72b: $0.352$ vs $0.183$ for Pearson correlation) is particularly illuminating as it mirrors how students actually approach multiple-choice questions. When students select answers, they typically compare options (A/B/C/D) simultaneously rather than evaluating each option's text in isolation. The index-based approach better captures this comparative decision-making process, while the text-based approach artificially forces sequential, independent evaluation of each option. This suggests that the stronger correlation in the index-based approach may stem from its better alignment with natural multiple-choice problem-solving strategies.

\begin{table}[t]
\centering
\caption{Analysis of LLM error patterns in multiple-choice questions: When LLMs answer incorrectly, do they select the same wrong answers as students? Results show remarkable alignment between LLM and student errors across model sizes (0.5B-72B parameters), with even small models selecting students' most common wrong answers over 50\% of the time. This surprising finding has important implications for educational technology: smaller models' tendency to make ``student-like'' mistakes, combined with their higher error rate, presents an innovative and cost-effective approach to generating pedagogically relevant distractors.}
\adjustbox{max width = \textwidth}{
\begin{tabular}{lccccc}
\toprule
\rowcolor{gray!21} \multicolumn{6}{c}{Index-based Likelihood} \\
& \# Incorrect & $1^{st}$ Dist (\%) & $2^{nd}$ Dist(\%) & $3^{rd}$ Dist(\%) & Alignment\\
\midrule
\textbf{qwen-0.5b} & 1781 & 51.6 & 32.6 & 15.8 & 0.757\\

\textbf{llama-1b} & 1887 & 49.2 & 33.1 & 17.7 & 0.731\\

\textbf{qwen-1.5b} & 1309 & 54.2 & 31.3 & 14.5 & 0.775\\

\textbf{llama-3b} & 1515 & 51.2 & 31.8 & 17.0 & 0.751\\

\textbf{qwen-3b} & 1078 & 55.9 & 28.8 & 15.2 & 0.784\\

\textbf{qwen-7b} & 794 & 56.5 & 28.1 & 15.4 & 0.785\\

\textbf{llama-8b} & 1259 & 51.4 & 29.7 & 18.9 & 0.748\\

\textbf{qwen-14b} & 561 & 58.6 & 29.1 & 12.3 & 0.803\\

\textbf{llama-70b} & 747 & 54.2 & 30.5 & 15.3 & 0.778\\

\textbf{qwen-72b} & 435 & 59.3 & 25.1 & 15.6 & 0.783\\
\midrule
\rowcolor{gray!21} \multicolumn{6}{c}{Text-based Likelihood} \\
& \# Incorrect & $1^{st}$ Dist (\%) & $2^{nd}$ Dist(\%) & $3^{rd}$ Dist(\%) & Alignment \\
\midrule
\textbf{qwen-0.5b} & 2116 & 39.0 & 31.9 & 29.1 & 0.648\\

\textbf{llama-1b} & 2079 & 38.7 & 32.2 & 29.1 & 0.647\\

\textbf{qwen-1.5b} & 1886 & 41.5 & 32.2 & 26.2 & 0.673\\

\textbf{llama-3b} & 1893 & 42.0 & 32.8 & 25.1 & 0.676\\

\textbf{qwen-3b} & 1758 & 41.9 & 32.3 & 25.9 & 0.670\\

\textbf{qwen-7b} & 1630 & 41.3 & 31.4 & 27.3 & 0.670\\

\textbf{llama-8b} & 1684 & 43.1 & 30.7 & 26.2 & 0.679\\

\textbf{qwen-14b} & 1451 & 42.9 & 30.7 & 26.5 & 0.682\\

\textbf{llama-70b} & 1352 & 43.7 & 29.8 & 26.5 & 0.687\\

\textbf{qwen-72b} & 1323 & 42.6 & 30.8 & 26.5 & 0.682\\

\bottomrule
\end{tabular}
}
\label{tab:rq2}
\end{table}

\subsubsection{Model Size Impact:} Figure \ref{img:rq1_alignment_size} reveals interesting scaling patterns across model sizes. For index-based approach, we observe a general upward trend in correlation as model size increases, with Qwen models showing particularly strong improvement from 0.5B to 14B parameters. The largest models maintain these strong correlations, demonstrating the benefits of model scale in capturing student reasoning patterns. The text-based approach shows more consistent improvement with scale across both model families, though with lower overall correlation values.

The figure also reveals that instruction-tuned variants (-instruct) of both model families generally show stronger correlations than their base counterparts, particularly noticeable in larger models. This suggests that instruction tuning may further enhance models' ability to capture student-like reasoning patterns. These findings provide quantitative evidence that LLMs' generation probabilities partially reflect student distractor preferences, particularly when the task format mirrors typical multiple-choice selection processes.

\subsection{RQ2: Alignment of LLM Mistakes and Student Misconception Patterns}

Our second research question examines whether LLMs, when answering incorrectly, tend to select the same distractors that commonly mislead students. To investigate this systematically, we analyzed incorrect answer selections from LLMs ranging from 0.5B to 72B parameters. For each LLM mistake, we tracked whether the model selected the distractor that was most commonly chosen by students ($1^{st}$ Dist), the second most common ($2^{nd}$ Dist), or the least common ($3^{rd}$ Dist). We tested this across two different prompting formats: index-based (where models select from A/B/C/D) and text-based (where models select the full answer text). The results in Table \ref{tab:rq2} reveal three key patterns:

\subsubsection{Similar Error Patterns Across Model Sizes:} Both small and large models show remarkably consistent patterns in selecting the first and second most common student distractors. For index-based prompting, the smallest model (Qwen-0.5B) selects the most common student distractor 51.6\% of the time and the second most common 32.6\% of the time, while the largest model (Qwen-72B) shows similar proportions at 59.3\% and 25.1\% respectively. This consistency suggests that the ability to capture student misconception patterns is not solely dependent on model size.

\subsubsection{A Cost-Effective Approach to Distractor Generation:} Our findings suggest a surprisingly cost-effective approach to automated distractor generation: using smaller language models. The key insight is that while smaller models make more mistakes, they make ``student-like'' mistakes. Specifically, when Qwen-0.5B ($n=1781$ incorrect answers) and Qwen-72B ($n=435$ incorrect answers) make mistakes, they show similar alignment with student error patterns, despite the dramatic difference in model size and overall accuracy. This opens up an efficient pathway for generating plausible distractors by:
\begin{enumerate}
    \item Leveraging smaller models' higher error rate to generate more potential distractors
    \item Using their natural alignment with student misconception patterns to ensure these distractors are pedagogically relevant
    \item Taking advantage of their computational efficiency and lower resource requirements
\end{enumerate}

\subsubsection{Impact of Likelihood Calculation Approach:} The method of presenting answer choices to LLMs significantly affects their alignment with student error patterns. Index-based prompting (using A/B/C/D) consistently shows higher alignment scores compared to text-based prompting across all model sizes. For example, Qwen-72B achieves a 59.3\% first-distractor selection rate with index-based prompting versus 42.6\% with text-based prompting, suggesting that simpler answer formats might better capture natural misconception patterns.

\section{Conclusion}
Our study provides the first quantitative framework for evaluating whether LLMs naturally capture patterns in how students select distractors in multiple-choice questions. Through our dual analysis framework, we found moderate correlations between LLM-assigned generation probabilities and student distractor selection distributions. These findings suggest that while LLMs show some alignment with student response patterns, they may not fully capture the cognitive processes underlying student choices. Interestingly, we found that smaller models, despite making more mistakes overall, show similar alignment with student error patterns as larger models. This suggests that computationally efficient smaller models could be valuable tools for distractor generation. Beyond this practical insight, the partial alignment between LLMs and student responses suggests an interesting opportunity: LLMs might generate distractors that probe different aspects of student understanding than traditional human-designed ones targeting known misconceptions. Future work should explore hybrid approaches that combine LLM capabilities with human expertise in educational assessment design, while remaining mindful that LLMs' underlying approach to problem-solving may fundamentally differ from human cognitive processes.

\section*{Acknowledgments}
This work was supported by NSF grant 1842378, ONR grant N0014-20-1-2534, AFOSR grant FA9550-22-1-0060, a Vannevar Bush Faculty Fellowship, OpenAI, and ONR grant N00014-18-1-2047.

\bibliographystyle{unsrt}  
\bibliography{custom,dsp}  

\appendix
\section{Statistical Correlation}
\label{app:stats-corr}
For each multiple-choice question $q$ with answer choices $A_q = \{a_1, a_2, \cdots, a_n \}$, we computed the correlation between student responses distribution $P_s(a_i \mid q)$ and language model likelihood $P_m(a_i \mid q)$ using the following statistical methods.

\subsection{Pearson's Correlation}
We use Pearson product-moment correlation~\cite{freedman2007statistics} coefficient $r_q$ to calculate the linear relationships between student performance distribution $P_s(a_i|q)$ and language model likelihood $P_m(a_i|q)$ . For a sample of $n$ questions, the coefficient is defined as:

$$r_q = \frac{\sum_{i=1}^n (P_s(a_i \mid q) - \mu_s)(P_m(a_i \mid q) - \mu_m)}{\sqrt{\sum_{i=1}^n (P_s(a_i \mid q) - \mu_s)^2}\sqrt{\sum_{i=1}^n (log P_m(a_i \mid q) - \mu_m)^2}}$$

where $\mu_s$ and $\mu_m$ denote the sample means. The coefficient $r_q \in [-1, 1]$ provides a normalized measure of linear dependence, where $|r_q| = 1$ indicates perfect linear correlation and $r_q = 0$ indicates no linear correlation. 

\subsection{Spearman's Rank Correlation}
We apply Spearman's rank correlation~\cite{zar2005spearman} coefficient $\rho_q$ to measure the non-linear monotonic relationships by operating on the ranks instead of raw values. Let $\text{rank}(P_s(a_i \mid q))$ and $\text{rank}(P_m(a_i \mid q))$ denote the rankings of the $i$-th answer choice when ordered by student response distribution and model likelihood respectively. The Spearman's rank coefficient is defined as:

$$\rho_q = 1 - \frac{6\sum_{i=1}^n d_i^2}{n(n^2-1)}$$

where $d_i = \text{rank}(P_s(a_i \mid q)) - \text{rank}(P_m(a_i \mid q))$ represents the rank difference for the $i$-th answer choice. The coefficient $\rho_q \in [-1, 1]$, where -1 indicates perfect negative correlation (inverse rankings), +1 indicates perfect positive correlation (identical rankings), and 0 suggests no correlations between the rankings.

\subsection{Kendall's tau Correlation}
We compute Kendall's tau~\cite{kendall1938new} correlation coefficient $\tau_q$ to measure the concordance between model predictions and student responses through pairwise rankings. For any two answer choices $i$ and $j$, a pair is considered concordant if their relative ordering is consistent across both distributions (i.e., if $P_s(a_i \mid q) > P_s(a_j \mid q)$ when $P_m(a_i \mid q) > P_m(a_j \mid q)$, or $P_s(a_i \mid q) < P_s(a_j \mid q)$ when $P_m(a_i \mid q) < P_m(a_j \mid q)$), and discordant if their ordering differs. Kendall's tau correlation is defined as:

$$\tau_q = \frac{2(n_c - n_d)}{n(n-1)}$$

where $n_c$ is the number of concordant pairs, $n_d$ is the number of discordant pairs. The coefficient $\tau_q \in [-1, 1]$, where values closer to ±1 indicate stronger alignment between model likelihoods and student answer distributions (positive for similar rankings, negative for opposite rankings), while 0 indicates no alignment.

\end{document}